\documentclass[10pt,twocolumn,letterpaper]{article}

\usepackage{wacv}
\usepackage{times}
\usepackage{epsfig}
\usepackage{booktabs}
\usepackage{multirow}
\usepackage{amsmath}
\usepackage{amssymb}
\usepackage{algorithm}
\usepackage[]{algpseudocode}
\usepackage{lineno,hyperref}

\wacvfinalcopy % *** Uncomment this line for the final submission

\begin{document}

%%%%%%%%% TITLE
\title{Apple Flower Detection using Deep Convolutional Networks}

\author{Philipe A. Dias\\
Marquette University\\
Electrical and Computer Engineering\\
Milwaukee, Wisconsin, USA \\
{\tt\small philipe.ambroziodias@marquette.edu}
\and
Amy Tabb\\
USDA-ARS-AFRS\\
Kearneysville, West Virginia, USA\\
{\tt\small amy.tabb@ars.usda.gov}
\and
Henry Medeiros\\
Marquette University\\
Electrical and Computer Engineering\\
Milwaukee, Wisconsin, USA \\
{\tt\small henry.medeiros@marquette.edu}
}

\maketitle
%\thispagestyle{empty}

%%%%%%%%% ABSTRACT
\begin{abstract}
To optimize fruit production, a portion of the flowers and fruitlets of apple trees must be removed early in the growing season. The proportion to be removed is determined by the bloom intensity, i.e., the number of flowers present in the orchard. Several automated computer vision systems have been proposed to estimate bloom intensity, but their overall performance is still far from satisfactory even in relatively controlled environments. With the goal of devising a technique for flower identification which is robust to clutter and to changes in illumination, this paper presents a method in which a pre-trained convolutional neural network is fine-tuned to become specially sensitive to flowers. 
Experimental results on a challenging dataset demonstrate that our method significantly outperforms three approaches that represent the state of the art in flower detection, with recall and precision rates higher than $90\%$. Moreover, a performance assessment on three additional datasets previously unseen by the network, which consist of different flower species and were acquired under different conditions, reveals that the proposed method highly surpasses baseline approaches in terms of generalization capability.\footnote{The citation information for this article is: P. A. Dias, A. Tabb, and H. Medeiros, “Apple flower detection using deep convolutional networks,” Computers in Industry, vol. 99, pp. 17 $-–$ 28, Aug. 2018.
DOI: 10.1016/j.compind.2018.03.010}, \footnote{Mention of trade names or commercial products in this publication is solely for the purpose of providing specific information and does not imply recommendation or endorsement by the U.S. Department of Agriculture. USDA is an equal opportunity provider and employer.}
\end{abstract}

\section{Introduction}
Various studies have established the relationships between bloom intensity, fruit load and fruit quality \cite{Forshey1986,Link2000}. Together with factors such as climate, bloom intensity is especially important to guide thinning, which consists of removing some flowers and fruitlets in the early growing season. Proper thinning directly impacts fruit market value, since it affects fruit size, coloration, taste and firmness. 

\begin{figure}[!t]
  \centering
  \includegraphics[scale=.8]{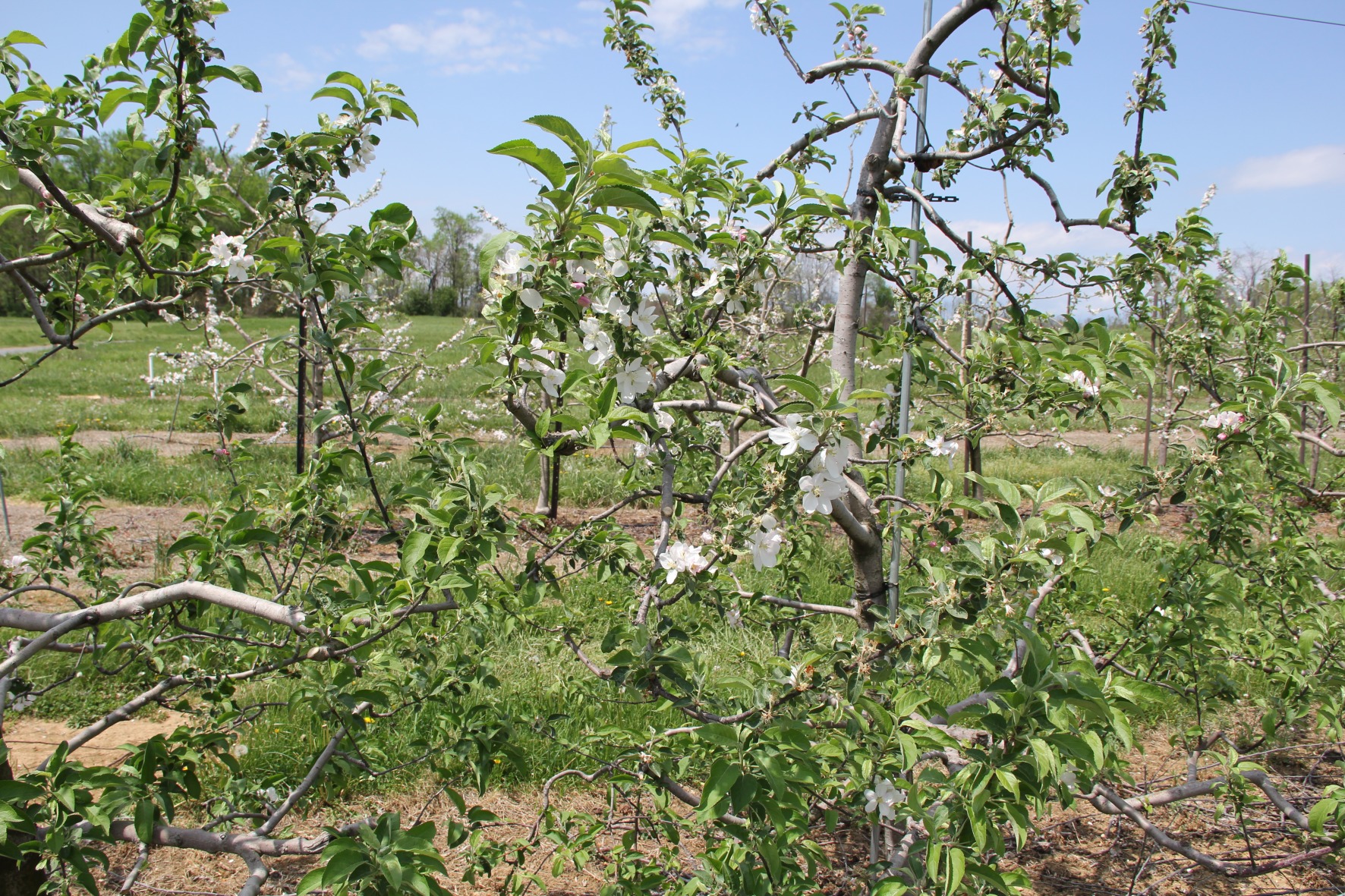}
  \caption{\textbf{Best viewed in color.} Example of image from a flower detection dataset used in this paper.
}
  \label{fig:bigexample}
\end{figure}

Despite its importance, there has been relatively limited progress so far in automating bloom intensity estimation. Currently, this activity is typically carried out manually with the assistance of rudimentary tools. More specifically, it is generally done by inspecting a random sample of trees within the orchard and then extrapolating the estimates obtained from individual trees to the remainder of the orchard \cite{Gongal2016}. As the example in Figure \ref{fig:bigexample} illustrates, obstacles that hamper this process are: 1) manual tree inspection is time-consuming and labor-intensive, which contributes to making labor responsible for more than $50\%$ of apple production costs \cite{Singh2010}; 2) estimation by visual inspection is characterized by large uncertainties and is prone to errors; 3) extrapolation of the results from the level of the inspected trees to the row or parcel level relies heavily on the grower's experience; and 4) inspection of a small number of trees does not provide information about the spatial variability which exists in the orchard, making it difficult to develop and adopt site-specific crop load management strategies that could lead to optimal fruit quality and yield.

With the goal of introducing more accurate and less labor intensive techniques for the estimation of bloom intensity, machine vision systems using different types of sensors and image processing techniques have been proposed \cite{Kapach2012}. Most existing methods, which are mainly based on simple color thresholding, have their applicability hindered especially by variable lighting conditions and occlusion by leaves, stems or other flowers \cite{Gongal2015}. 

Inspired by successful works using convolutional neural networks (CNNs) in multiple computer vision tasks, we propose a novel method for apple flower detection based on features extracted using a CNN. In our approach, an existing CNN trained for saliency detection is fine-tuned to become particularly sensitive to flowers. This network is then used to extract features from portraits generated by means of superpixel segmentation. After dimensionality reduction, these features are fed into a pre-trained classifier that ultimately determines whether each image region contains flowers or not. The proposed method significantly outperformed state-of-the-art approaches on four datasets composed of images acquired under different conditions.

Our main contributions are: 
\begin{enumerate}
\item[1)] a novel CNN-based flower detection algorithm;
\item[2)] an extensive evaluation on a challenging dataset acquired under realistic and uncontrolled conditions;
\item[3)] an analysis of the generalization capability of the proposed approach on additional datasets previously unseen by the evaluated models.
\end{enumerate}

The remainder of paper is organized as follows. Section \ref{sec:relatedwork} discusses the most relevant existing approaches for automated flower and fruit detection. Our proposed approach is described in Section \ref{sec:methods}, which also includes a description of three baseline comparison methods as well as implementation details. Experiments performed to evaluate the impact of specific design choices are described in Section \ref{sec:results}, followed by an extensive comparison of our optimal model against the baseline methods on four different datasets. Our concluding remarks are presented in Section \ref{sec:conclusion}. 

\section{Related Work}
\label{sec:relatedwork}
While existing techniques employed for flower detection are based only on color information, methods available for fruit quantification exploit more modern computer vision techniques. For this reason, in this section we first review the most relevant works on automated flower detection, followed by a discussion of the relevant literature on fruit quantification. Moreover, to make this article self-contained and therefore accessible to a wider audience, we also provide a brief introduction to the fundamentals of CNNs.

\subsection{Flower and fruits quantification}
Aggelopoulou and colleagues presented in \cite{Aggelopoulou2011} one of the first works using computer vision techniques to detect flowers. That method is based on color thresholding and requires image acquisition at specific daylight times, with the presence of a black cloth screen behind the trees. Thus, although its reported error in predicted yield is relatively low ($18\%$), such approach is applicable only for that controlled scenario.

Similar to the work of Thorp and Dierig ~\cite{Thorp2011} for identification of \textit{Lesquerella} flowers, the technique described by Ho\v{c}evar et al. in \cite{Hocevar2014} does not require a background screen, but it is still not robust to changes in the environment. The image analysis procedure is based on hard thresholding according to color (in the HSL color space) and size features, such that parameters have to be adjusted whenever changes in illumination (daylight/night), in flowering density (high/low concentration) or in camera position (far/near trees) occur.

Horton and his team described in \cite{Horton2017} a system for peach bloom intensity estimation that uses a different imaging approach. Based on the premise that the photosynthetic activity of this species increases during bloom period, the system relies on multispectral aerial images of the orchard, yielding an average detection rate of $84.3\%$ for $20$ test images. Similarly to the aforementioned methods, the applicability of this method also has the intrinsic limitation of considering only color/spectral information (thresholding near-infrared and blue bands), such that its performance is sensitive to changes in illumination conditions. 

More advanced computer vision techniques have been employed for fruit quantification \cite{Kapach2012}. A multi-class image segmentation for agrovision is proposed by Hung et al. in \cite{Hung2013}, classifying image pixels into leaves, almonds, trunk, ground and sky. Their method combines sparse autoencoders for feature extraction, logistic regression for label associations and conditional random fields to model correlations between pixels. Some other methods are based on support vector machine (SVM) classifiers that use information obtained from different shape descriptors and color spaces as input \cite{Das2015,Ji2012}. Compared to existing methods for flower detection, these methods are more robust since morphological characteristics are taken into account. As many other shape-based and spectral-based approaches \cite{Linker2012,Wachs2010,Wang2012,Dorj2017}, these techniques are, however, still limited by background clutter and variable lighting conditions in orchards \cite{Gongal2016}. 

Recent works on fruit quantification include the use of metadata information. Bargoti and colleagues in \cite{Bargoti2016} built on \cite{Hung2013} to propose an approach that considers pixel positions, orchard row numbers and the position of the sun relative to the camera. Similarly, Cheng et al. \cite{Cheng2017} proposed the use of information such as fruit number, fruit area, area of apple clusters and foliage area to improve accuracy of early yield prediction, especially in scenarios with significant occlusion. However, the inclusion of metadata is highly prone to overfitting, particularly when limited training data is available and the variability of the training set is hence low \cite{Bargoti2016}.

\subsection{Deep learning}
Following the success of Krizhevsky's model \cite{Krizhevsky2012} in the ImageNet 2012 Challenge, deep learning methods based on CNNs became the dominant approach in many computer vision tasks. The architecture of traditional CNNs consists of a fixed-size input, multiple convolutional layers, pooling (downsampling) layers and fully connected layers \cite{Guo2016}. Winner of the ImageNet 2013 Classification task, the Clarifai model is one such network \cite{Zeiler2014}. Illustrated on the right side of Figure \ref{fig:flow}, it takes input image portraits of size $227 \times 227$ pixels, which traverse a composition of $5$ convolutional layers (C1-C5) and $3$ fully connected layers (FC6-FC7 and the softmax FC8). Each type of layer plays a different role within the CNN architecture: while convolutional layers allow feature extraction, the latter fully connected layers act on this information to perform classification.

In computer vision and image processing, a \textit{feature} corresponds to information that is meaningful for describing an image and its regions of interest for further processing. Feature extraction is therefore crucial in image analysis, since it represents the transition from pictorial (qualitative) to nonpictorial (quantitative) data representation \cite{Marques2011}. Rather than relying on hand-engineered features (e.g. HOG \cite{Dalal2005Histograms}), deep CNNs combine multiple convolutional layers and downsampling techniques to learn hierarchical features, which are a key factor for the success of these models \cite{LeCun2015}. As described in \cite{Zeiler2014}, the convolutional layers C1-C2 learn to identify low-level features such as corners and other edge/color combinations. The following layers C3-C5 combine this low-level information into more complex structures, such as motifs, object parts and finally entire objects.

Traditional deep CNNs are composed of millions of learned parameters (over $60$ million in AlexNet \cite{Krizhevsky2012}), such that large amounts of labeled data are required for network training. Deep learning models became feasible relatively recently, after the introduction of large publicly available datasets, of graphics processing units (GPUs), and of training algorithms that exploit GPUs to efficiently handle large amounts of data \cite{LeCun2015}. Nevertheless, gathering domain specific training data is an expensive task. One alternative to reduce the required amount of labeled data is data augmentation, a technique proven to benefit the training of multiple machine learning models \cite{Wong2016}. It is typically performed by applying transformations such as translation, rotation and color space shifts to pre-labeled data. 

In addition, transfer learning approaches such as fine-tuning have been investigated. Earlier layers of a deep network tend to contain more generic information (low-level features), which are then combined by the latter layers into task specific objects of interest. Thus, a network that can recognize different objects present in a large dataset must contain a set of low-level descriptors robust enough to characterize a wide range of patterns. Under this premise, fine-tuning procedures typically aim at adjusting the higher-level part of a network pre-trained on a large generic dataset, rather than training the full network from scratch. This greatly reduces the need for task-specific data, since only a smaller set of parameters has to be refined for the particular application \cite{Yosinski2014}. 

At the classification side, most CNN architectures employ fully connected layers for final categorization. They determine which features are mostly correlated to each specific class employing a logistic regression classifier. For scenarios in which the output is binary, consistent albeit small improvements on popular datasets have been demonstrated by replacing the final CNN layer by a SVM classifier \cite{Tang2013}. SVM models tend to generalize better than logistic regression, since they target a solution that not only minimizes the training error, but also maximizes the margin distance between classes. 

Following the success of CNNs on image classification tasks, the work of Girschick et al. \cite{Girshick2014} introduced the concept of region-based CNNs (R-CNN), outperforming by a large margin previous hand-engineered methods for object detection. In that work, a CNN is first pre-trained on a large auxiliary dataset (ImageNet) and then fine-tuned using a smaller but more specific dataset (PASCAL dataset for object detection). The Faster R-CNN proposed in \cite{Ren2015} improved this model by replacing selective search \cite{Uijlings2013} with the concept of Region Proposal Network (RPN), which shares convolutional layers with the classification network. Both modules compose a single, unified network for object detection. 

Recent works adapt the Faster R-CNN for fruit detection. Bargoti and Underwood in \cite{Bargoti2016b} present a Faster R-CNN trained for detection of mangoes, almonds and apples fruits on trees. Stein et al. in \cite{Stein2016} extended this model for tracking and localization of mangoes, combining it with a monocular multi-view tracking module that relies on a GPS system. Sa et al. in \cite{Sa2016_DeepFruits} applied the Faster R-CNN to RGB and near-infrared multi-modal images. Each modality was fine-tuned independently, with optimal results obtained using a late fusion approach. Still in the context of agricultural applications, CNNs have been also successfully used for plant identification from leaf vein patterns \cite{Grinblat2016}. 

In summary, existing methods for flower identification are based on hand-engineered image processing techniques that work only under specific conditions. Color and size thresholding parameters composing these algorithms have to be readjusted in case of variations of lightning conditions, camera position with respect to the orchard (distance and angle), or expected bloom intensity. Recent techniques employed for fruit quantification exploit additional features and machine learning strategies, providing insights to further develop strategies for flower detection. Aiming at a technique for flower identification that is robust to clutter, changes in illumination and applicable for different flower species, we therefore propose a novel method in which an existing CNN trained for saliency detection is fine-tuned to become particularly sensitive to flowers. 

\section{Proposed Approach}
\label{sec:methods}

In this section, we first describe the prediction steps performed by our method, i.e., the sequence of operations applied to an image in order to detect the presence of flowers. Subsequently, we describe the fine-tuning procedure carried out to obtain the core component of our model: a CNN highly sensitive to flowers. We conclude with a discussion of alternative flower detection approaches against which we evaluate our proposed method and a brief mention of relevant details regarding the implementation of our methods. 

\begin{figure*}[t]
  \centering
  \includegraphics[width=\textwidth]{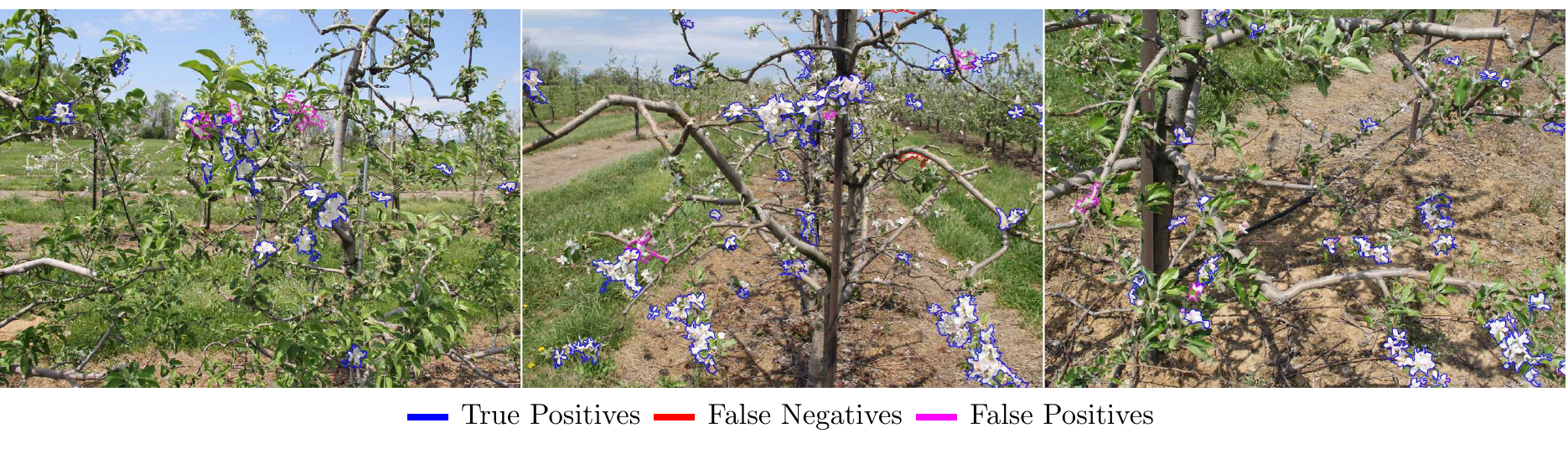}
    \caption{\textbf{Best viewed in color.} Diagram illustrating the sequence of image analysis tasks performed by the proposed model for flower identification. Layers FC7-FC8 are used only during fine-tuning (training). For final prediction, features are collected from the output of layer FC6. Each task and its corresponding output (shown above the arrows) are described in Algorithm \ref{alg:pseudocode}. }
    \label{fig:flow}
\end{figure*}

In the discussion that follows, we will refer to our proposed approach for flower detection as the \emph{CNN+SVM} method. As illustrated in Figure \ref{fig:flow}, our CNN+SVM method consists of three main steps: i) computation of region proposals; ii) feature extraction using our fine-tuned CNN, which follows the Clarifai architecture \cite{Zeiler2014}; and iii) final classification of each region according to the presence of flowers. The operations that comprise these steps are described in detail below. In our description, we make reference to Algorithm \ref{alg:pseudocode}, which lists the operations performed by our method on each input image. The sensitivity of the method to specific design choices is detailed in Section \ref{sub:design}.

\textit{1) Step 1 - Region proposals:} The first step in the proposed method consists of generating region proposals by grouping similar nearby pixels into superpixels, which are perceptually meaningful clusters of variable size and shape (Line \ref{ln:seg} of Algorithm \ref{alg:pseudocode}). To this end, we use the \textit{simple linear iterative clustering} (SLIC) superpixel algorithm. Currently one of the most widely-used algorithms for superpixel segmentation, it adapts \textit{k}-means clustering to group pixels according to a weighted distance measure that considers both color and spatial proximity \cite{Achanta2012}. For additional information on superpixel approaches, we refer the reader to the review provided in \cite{Stutz2016}. The second leftmost image in Figure \ref{fig:flow} illustrates the superpixels $s_i \in \mathbb{S}$ generated by the SLIC algorithm when applied to a typical image obtained in an orchard. 

Although other approaches such as the Faster R-CNN \cite{Ren2015} provide a unified architecture in which both region proposal and classification modules can be fine-tuned for a specific task, they have more parameters that need to be learned in a supervised manner. Since in most cases flowers are salient with respect to its surrounding background, an unsupervised, local-context based approach such as superpixel segmentation should be sufficient to obtain region proposals suitable for flower detection. 

\begin{algorithm}
  \caption{Proposed approach for flower detection}\label{euclid}
  \label{alg:pseudocode}
  \begin{algorithmic}[1]
      \Require{Image $I$.}
      \Ensure{Regions in $I$ containing flowers.} % output
      \State Segment $I$ into set of superpixels $\mathbb{S}$ using SLIC. \label{ln:seg}
      \For {each superpixel $s_i \in \mathbb{S}$}
      \State Crop smallest squared portrait $p_i$ enclosing $s_i$. \label{ln:crop}
      %\State \smash{\parbox[t]{\dimexpr\linewidth-\algorithmicindent}{Generate $\hat{p_i}$ by mean-padding the background surrounding $s_i$ in $p_i$.}}\label{ln:pad} 
   	\State{Generate $\hat{p_i}$ by mean-padding the background surrounding $s_i$ in $p_i$.}\label{ln:pad} 
      %\State \smash{\parbox[t]{\dimexpr\linewidth-\algorithmicindent}{Extract features ${f}_i$ from the mean-centered $\hat{p_i}$ using the fine-tuned CNN.}}\label{ln:feat}
	\State {Extract features ${f}_i$ from the mean-centered $\hat{p_i}$ using the fine-tuned CNN.}\label{ln:feat}      
	\State Obtain $\hat{f}_i$ by performing PCA analysis on ${f}_i$. \label{ln:pca}
      \State Classify $s_i$ by applying pre-trained SVM on $\hat{{f}}_i$. \label{ln:svm}
      \EndFor
  \end{algorithmic}
\end{algorithm}

Once the image is segmented into superpixels, as Algorithm \ref{alg:pseudocode} indicates, we iterate over each superpixel in the image. Since the input size required by the Clarifai CNN model is $227 \times 227$, we first extract the smallest square portrait enclosing the superpixel under analysis (Line \ref{ln:crop}), which we denote $p_i$. The output of this step is illustrated in the third leftmost image of Figure \ref{fig:flow} for one superpixel. The background surrounding the superpixel of interest within a portrait is then padded with the training set mean, i.e. the average RGB color of all images composing the dataset (greenish color). Finally, the portrait is resized to $227 \times 227$ (Line \ref{ln:pad}). The resulting region proposal, $\hat{p}_i$, is illustrated in the fourth image of Figure \ref{fig:flow}.

\textit{2) Step 2 - Feature extraction:} In the feature extraction step (Line \ref{ln:feat}), each of the portraits generated above is mean-centered and then evaluated individually by our CNN. The mean-centering step consists of subtracting from the portrait the same average training set RGB mean used for padding its background. This procedure is commonly employed to facilitate training convergence of deep learning models, since it ensures similarly ranged features within the network. For each input portrait, we collect as features the output of the rectified linear unit (ReLU) associated with the first fully connected layer (FC6). With a dimensionality of $N=4,096$, the feature vector $f_i \in \mathbb{R}^N$ collected at this stage of the network encapsulates the hierarchical features extracted by layers $C1-C5$, which contain the key information required for accurate classification. 

\textit{3) Step 3 - Classification:} To classify each proposed region as containing a flower or not, we first perform principal component analysis (PCA) to reduce the feature dimensionality to a value $k<N$ such that the new feature vector $\hat{f}_i \in \mathbb{R}^k$ (Line \ref{ln:pca}). As demonstrated in our experimental evaluation in Section \ref{ss:dimensionality} a value of $k=69$, which corresponds to approximately $94\%$ of the original variance of the data, provides performance levels virtually identical to those of the original features. Finally, based on these features a pre-trained SVM model binary classifies superpixels according to the presence of flowers (Line \ref{ln:svm}). Details on SVM training are provided in the next section.

\begin{figure*}[t]
  \centering
  \includegraphics[width=\textwidth]{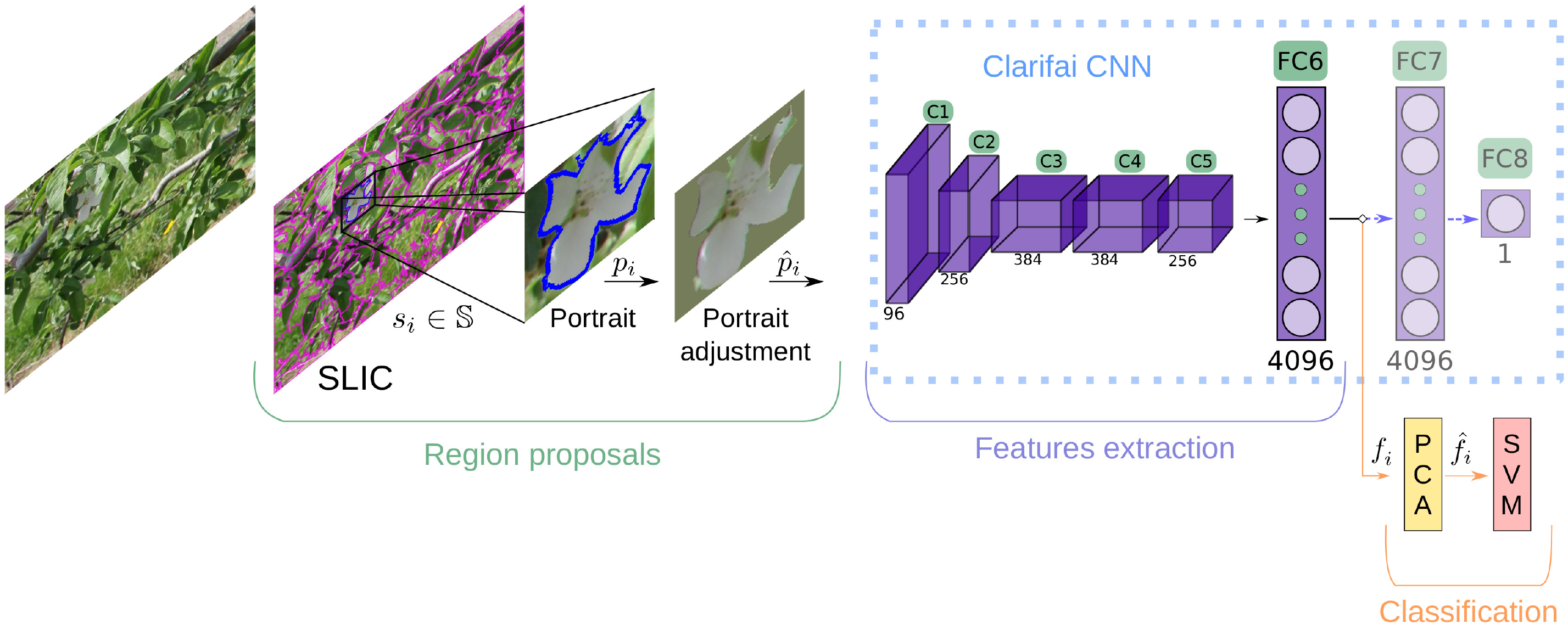}
    \caption{\textbf{Best viewed in color.} Examples of images composing the \textit{AppleA} dataset, with the corresponding detection provided by the proposed algorithm.\protect\footnotemark[1]}    
    \label{fig:imgsamples}
\end{figure*}

\footnotetext[1]{More examples are available in the supplementary material.}

\subsection{Network fine-tuning and SVM training}
\label{sub:tuning}
Based on the techniques introduced by Girshick et al. in \cite{Girshick2014} and Zhao et al. in \cite{Zhao2015} for object and saliency detection, in our model an existing CNN architecture is made particularly sensitive to flowers by means of fine-tuning. In the work of Zhao et al. \cite{Zhao2015}, the Clarifai model \cite{Zeiler2014} was adopted as the starting point and fine-tuned for saliency detection. We further tuned Zhao et al.'s model for flower identification using labeled portraits from our training set, which we describe below.

The generation of training samples for network tuning takes place similarly to prediction. For each labeled image composing the training set, we computed region proposals according to Step 1 described above. Using these training examples, $10,000$ backpropagation training iterations were performed in order to minimize the network classification error. After fine-tuning, we computed the CNN features of the training examples, reduced their dimensionality to $k=69$, and used them to train the SVM classifier.

% for the final version, include further details of orchard location (removed for blind review process)
\vspace{10pt}
\paragraph{Image dataset} Images of apple trees were collected using a camera model Canon EOS 60D under natural daylight illumination (i.e. uncontrolled environment). This dataset, which we refer to as \textit{AppleA}, is composed of a total of $147$ images with resolution of $5184 \times 3456$ pixels acquired under multiple angles and distances of capture. Figure \ref{fig:imgsamples} shows some images that comprise this dataset. For performance evaluation and learning purposes, the entire dataset was labeled using a MATLAB GUI in which the user selected only superpixels that contain parts of flowers in approximately half of its total area. As summarized in Table \ref{tab:labels}, the labeled images were randomly split into training and validation sets composed of $100$ and $47$ images, respectively. This corresponds to a total of $91,488$ training portraits (i.e. superpixels) and $42,430$ validation ones. The training examples were used to fine-tune the network and train the SVM as described above. The validation examples were used in the performance evaluation discussed in Sections \ref{sub:design} and \ref{sub:comparison}.

\begin{table}[h]\centering
\setlength{\tabcolsep}{0.4em} 
  \caption{Statistics of the training and validation dataset (\textit{AppleA}).}
  \label{tab:labels}
  \begin{tabular}{@{}lcccc@{}}\toprule
	      &   & \multicolumn{3}{c}{\textbf{Portraits (i.e. superpixels)}}      \\ \cmidrule{3-5} 
	      & \multicolumn{1}{c}{\textbf{Images}} & {Positives} & {Negatives} & {Total} \\ \midrule[0.3pt]\midrule
\multicolumn{1}{l}{Training}   & $100$ & $3,691$ ($4\%$) & $87,797$ ($96\%$)  & $91,488$  \\ 
\multicolumn{1}{l}{Validation} & $47$  & $1,719$ ($4\%$) & $40,711$ ($96\%$)  & $42,430$  \\ 
\multicolumn{1}{l}{\textit{Total}}   	& $147$ & $5,410$ ($4\%$) & $128,508$ ($96\%$) & $133,918$ \\ 
    \bottomrule
  \end{tabular}
\end{table}

\begin{figure}[!ht]
  \centering
  \includegraphics[scale=1]{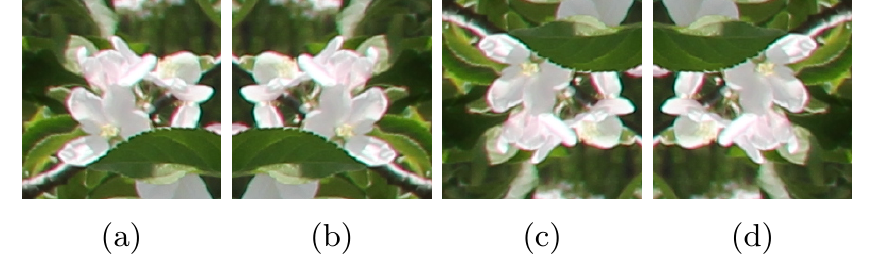}
 \caption{\textbf{Best viewed in color.} Example of data augmentation. a) Original portrait. b) Portrait mirrored with respect to the vertical axis, c) the horizontal axis, d) and  both axes.}
 \label{fig:augmentation}
\end{figure}

\vspace{10pt}
\paragraph{Data augmentation} According to our labeling, only $4\%$ of the samples contain flowers (positives). Imbalanced datasets represent a problem for supervised machine learning approaches, since overall accuracy measures become biased towards recognizing mostly the majority class \cite{Visa2005}. In our case, that means the learner would present a bias towards classifying the portraits as negatives. To overcome this situation and increase the amount of training data, we quadrupled the number of positive samples using data augmentation. As illustrated in Figure \ref{fig:augmentation}, this was accomplished by mirroring each positive sample with respect to: (i) the vertical axis, (ii) the horizontal axis, and (iii) both axes.

\vspace{10pt}
\paragraph{Parameters' optimization} Support vector machines are supervised learning models that search for a hyperplane that maximizes the margin distance to each class. This characteristic allows SVM models to generalize better than classifiers such as the ones based on logistic regression. For non-linearly separable data, kernel functions such as the popular radial basis function (RBF, or \textit{Gaussian}) are used. We refer to \cite{BenHur2010,Hsu2008} for further details on the formulation of SVMs. 

Two main parameters control the performance of SVMs with a Gaussian kernel function, the regularization cost $C$ and the width of the Gaussian kernel $\gamma$. By regulating the penalty applied to misclassifications, the parameter $C$ controls the trade-off between maximizing the margin with which two classes are separated and the complexity of the separating hyperplane. The parameter $\gamma$ regulates the flexibility of the classifier's hyperplane. For both parameters, excessively large values can lead to overfitting.

The optimization of $C$ and $\gamma$ is a problem without straightforward numerical solution. Therefore, it is typically solved using grid search strategies \cite{BenHur2010,Hsu2008} in which multiple parameter combinations are evaluated according to a performance metric. We adopt this strategy in this work.

\subsection{Comparison Approaches}
\label{sub:comparison_description}

As has been noted in Section \ref{sec:relatedwork}, current algorithms for automated identification of flowers are mostly based on binarization by thresholding information from different color-spaces (typically RGB or HSV) \cite{Aggelopoulou2011,Thorp2011}, occasionally combined with size filtering \cite{Hocevar2014}. We implement three alternative baseline approaches which reflect the state of the art in fruit/flower detection. The first implementation, which we call the \emph{HSV} method, replicates the algorithm described by Hocevar and his team in \cite{Hocevar2014}. Images are binarized at pixel-level based on HSV color information, followed by filtering according to minimum and maximum cluster sizes. 

We refer to the second baseline implementation as \emph{HSV+Bh}. Similar to our proposed approach, the starting point for this method is the generation of superpixels using the SLIC algorithm. We then compute a $100$-bin histogram of each superpixel in the HSV color space, which has the advantage of dissociating brightness from chromaticity and saturation. Studies on human vision and color-based image retrieval have demonstrated that most of the color information is contained in the hue channel, with the saturation playing a significant role in applications where identifying white (or black) objects is important \cite{Gonzalez2006,Stricker95hist}. In our experiments, we construct a single 1-D histogram consisting of $100$ bins, which corresponds to the concatenation of a 50-bin hue channel histogram, a 40-bin saturation histogram and a 10-bin value histogram. Afterwards, we use the Bhattacharyya distance \cite{Bhattacharyya1943} to compare each superpixel histogram against the histograms of all positive samples composing the training set. We compute the Bhattacharyya distance using a Gaussian kernel function, as formulated in \cite{Hoak2017,Hoseinnezhad2012}. The average Bhattacharyya distance is taken as the likelihood that the superpixel includes a flower, and superpixels with distance lower than an optimal threshold are classified as flowers.

Since the technique described above is based on the average Bhattacharyya distance in the HSV color space, it makes no distinction between poorly and highly informative training sample features. Its ability to make accurate classification decisions is therefore limited in such complex feature spaces. Inspired by works on fruit quantification \cite{Das2015,Ji2012}, we extend this method by combining the same HSV histograms with an SVM classifier for apple flower detection. That is, rather than determining whether a superpixel contains a flower based on the Bhattacharyya distance, we train an SVM classifier that uses the HSV histograms as inputs. We call this approach the \emph{HSV+SVM} method.

\subsection{Implementation Details}
\label{sec:paropt}
Most image analysis tasks were performed using MATLAB R2016b. Additionally, we used the open source Caffe framework \cite{Jia2014a} for fine-tuning and extracting features from the CNN. To reduce computation times by exploiting GPUs, we used the cuSVM software package for SVM training and prediction \cite{Carpenter2009}.

\section{Experiments and Results}
\label{sec:results}

Experiments were performed with three main goals. Our optimal CNN+SVM model extracts features from the CNN's first fully connected layer ($FC6$), reduces feature dimensionality to $69$, and performs final classification using SVM. Thus, we first evaluated the impact of these specific design choices on the final performance of our method.  We then compared it against the three baseline methods (HSV, HSV+Bh and HSV+SVM). Finally, we evaluated the performance of the proposed approach on previously unseen datasets to determine its generalization capability. 

As described in Section \ref{sub:tuning}, our datasets are severely imbalanced. In such scenarios, evaluations of performance using only accuracy measurements and ROC curves may be misleading, since they are insensitive to changes in the rate of class distribution. We therefore perform our analysis in terms of precision-recall curves (PR) and the corresponding $F_1$ score \cite{Fawcett2006}. Precision is normalized by the number of positives rather than the number of true negatives, so that false positive detections have the same relative weight as true positives. While the maximum $F_1$ score indicates the optimal performance of a classifier, the area under the respective PR curve (AUC-PR) corresponds to its expected performance across a range of decision thresholds, such that a model with higher AUC-PR is more likely to generalize better.

\subsection{Analysis of design choices}
\label{sub:design}
In order to validate our design choices, we performed experiments to evaluate how the final performance of the classifier is affected by: (i) the dimensionality of the feature space; (ii) the point where features are collected from the CNN; and (iii) the type of input portrait.

\subsubsection{Dimensionality analysis}
\label{ss:dimensionality}

PCA is one of the most widespread techniques for dimensionality reduction. It consists of projecting $N$-dimensional input data onto a $k$-dimensional subspace in such a way that this projection minimizes the reconstruction error (i.e. $L_2$ norm between original and projected data) \cite{Mohri2012}. PCA can be performed by computing the eigenvectors and eigenvalues of the covariance matrix and ranking principal components according to the obtained eigenvalues \cite{Smith2002}.

In this application, the original dimensionality corresponds to the number of elements in the CNN vectors extracted from a fully connected layer, i.e., $N=4,096$ as represented for the two last layers in Figure \ref{fig:flow}. The first two columns of Table \ref{tab:pcaf} show the reduced dimensionality $k$ of the feature vector and the corresponding ratio of the total variance of the $N$-dimensional dataset that is retained at that dimensionality for layer $FC6$. As the table indicates, the first most significant dimension alone already covers almost half of the total variance, and $23$ dimensions are sufficient to cover nearly $90\%$ of it.  

\begin{table}[!ht]\centering
  \setlength{\tabcolsep}{0.3em}
  \caption{Classification performance according to the number of principal components (dimensions) selected after applying PCA to the extracted features. Best results in terms of $F_1$ and AUC-PR are shown in boldface. }
  \label{tab:pcaf}
  \begin{tabular}{@{}lccccc@{}}\toprule
    \multicolumn{1}{c}{\shortstack[c]{\textbf{No. of} \\ \textbf{dimensions}}} & \multicolumn{1}{c}{\shortstack[c]{\textbf{Variance} \\ \textbf{ratio}}}  & \multicolumn{1}{c}{$\mathbf{F_1}$} & \multicolumn{1}{c}{\textbf{Recall}} & \multicolumn{1}{c}{\textbf{Precision}} & \multicolumn{1}{c}{\textbf{AUC-PR}}\\ \midrule[0.3pt]\midrule
    $1   $ & $48.3\%$ & $90.4\%$ & $92.2\%$ & $88.6\%$ & $96.5\%$ \\ 
    $2   $ & $63.7\%$ & $91.4\%$ & $92.7\%$ & $90.2\%$ & $94.0\%$ \\ 
    $5   $ & $79.9\%$ & $91.9\%$ & $92.3\%$ & $91.5\%$ & $96.9\%$ \\ 
    $15  $ & $87.4\%$ & $91.5\%$ & $92.6\%$ & $90.4\%$ & $94.3\%$ \\ 
    $23  $ & $89.6\%$ & $\mathbf{92.1\%}$ & $92.9\%$ & $91.2\%$ & $95.2\%$ \\ 
    $69  $ & $93.8\%$ & $91.9\%$ & $92.6\%$ & $91.2\%$ & $\mathbf{97.2}\%$ \\ 
    $150 $ & $95.8\%$ & $91.3\%$ & $92.7\%$ & $90.0\%$ & $97.1\%$ \\ 
    $300 $ & $97.2\%$ & $91.6\%$ & $91.6\%$ & $91.7\%$ & $97.2\%$ \\ 
    $500 $ & $98.0\%$ & $91.8\%$ & $91.8\%$ & $91.8\%$ & $95.0\%$ \\ 
    $1080$ & $99.0\%$ & $91.7\%$ & $91.5\%$ & $91.8\%$ & $94.9\%$ \\  
    \bottomrule
  \end{tabular}
\end{table}

\begin{figure*}[h]
	\centering	      
	\includegraphics[width=\textwidth]{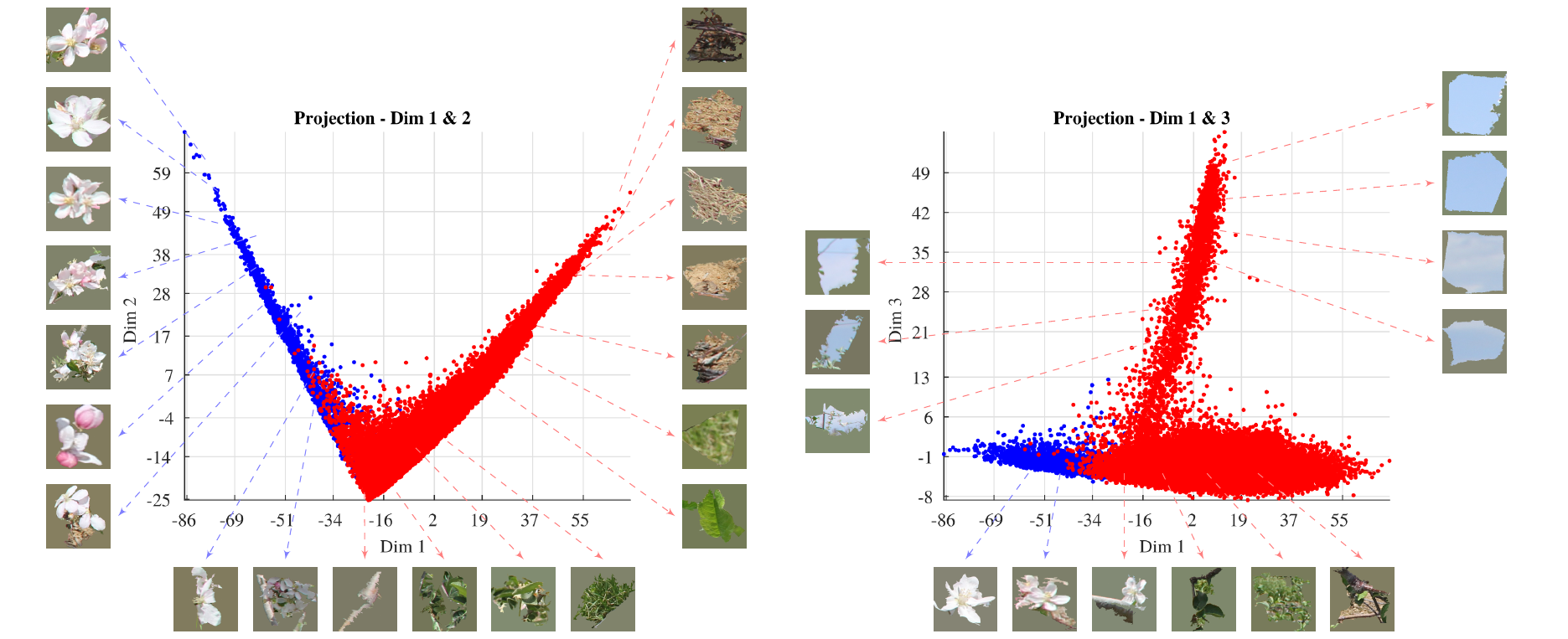}
	\caption{\textbf{Best viewed in color.} Projections of samples on 2D feature spaces, with positive samples in blue and negatives ones in red. \textit{Left:} sample distribution on the plane corresponding to dimensions $1$ and $2$ according to PCA. \textit{Right:} sample distribution on the plane corresponding to dimensions $1$ and $3$.} 
  \label{fig:proj13Ex}  
\end{figure*}

We investigated then how samples are mapped into the lower dimensional feature space. Figure \ref{fig:proj13Ex} shows the projections in dimensions $1$ and $2$ as well as dimensions $1$ and $3$. These plots illustrate how the convolutional network maps the samples into a space where it is possible to differentiate between multiple clusters. With \textit{dim.} as an abbreviation for \textit{dimension}, let $\downarrow$ denote low dimensionality values and $\uparrow$ high values, respectively. The following clusters can be identified: flowers ($\downarrow$ dim.1, $\uparrow$ dim.2); grass/floor ($\uparrow$ dim.1, $\uparrow$ dim.2); branches/leaves ($\downarrow$ dim.2); sky ($\uparrow$ dim.3). This indicates that positive and negative samples are almost linearly separable even for $2D$ projections of the original feature space.

To quantitatively assess how the classification performance is affected by the dimensionality of the feature space, we trained SVM classifiers for different numbers of dimensions. For each dimensionality, Table \ref{tab:pcaf} presents the optimal performance metrics and corresponding AUC-PR. As expected, these results demonstrate that the impact of dimensionality on the optimal performance of our method is rather low. A very good performance is already obtained using a 2D feature space, with both $F_1$ score and AUC-PR only around $0.7\%$ and $3.2\%$ lower than the highest obtained values, respectively. In terms of optimal recall and precision, this is equivalent to missing extra $4$ positive samples out of $1,719$, while including more $19$ false-positives out of $40,711$. Moreover, the table shows that a dimensionality of $69$ is nearly optimal: the performance in terms of optimal $F_1$ score is only $0.2\%$ lower than the highest obtained value ($23$ dimensions) and it is optimal in terms of AUC-PR. 

Although in the discussion above we present results obtained using SVMs, such a high separability even for low dimensionalities indicates that the final prediction accuracy of our model is almost independent of the type of classifier employed. This conjecture is validated in the next subsection, where we demonstrate that the performance of our system does not change significantly by either including an additional fully connected layer to our CNN or by carrying out classification using the using network's softmax layer directly.

\subsubsection{Feature analysis}
\begin{figure}[!h]
	\centering	      
	\includegraphics[scale=1]{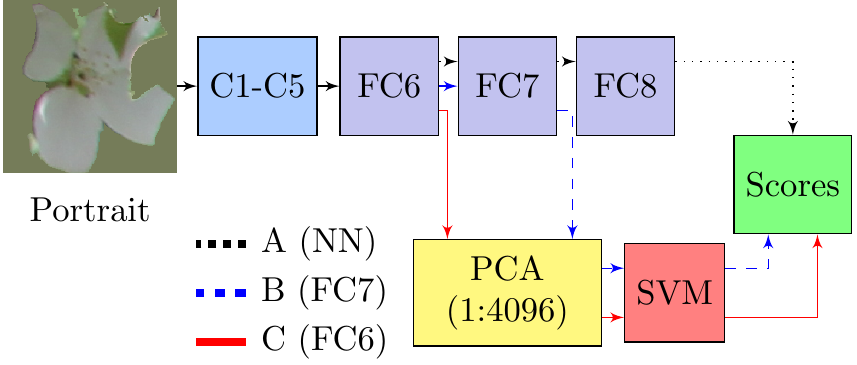}
  \caption{Diagram illustrating how classification scores are computed using the extracted features.}
  \label{fig:score}
\end{figure}

As explained in Section \ref{sec:methods}, after fine-tuning the model, we use it to extract features that allow the classification of superpixels according to the presence of flowers within them. Three combinations of features and classification mechanisms were investigated: (A) predict using solely the neural network, by means of its softmax output layer; (B) train a SVM classifier on features collected after the last fully connected layer (FC7); (C) train a SVM classifier on features collected after the first fully connected layer (FC6). Figure \ref{fig:score} shows the points where features are collected and how classification scores are computed using these features. Following the notation used in Figure \ref{fig:flow}, C1-C5 correspond to the convolutional layers of the fine-tuned Clarifai network, FC6-FC7 are the fully connected layers, and FC8 is the softmax layer.

For approaches B and C, features are collected from the output of the rectified linear units (ReLUs) located right after the respective fully connected layers. The same sequence of operations is performed for both methods B and C, i.e., the framework is the same regardless of whether the features are collected from the last (FC7) or first fully connected layer (FC6). Based on the results obtained in the previous section, for both cases $69$ dimensions are kept after PCA analysis.

\begin{figure}[!ht]
	\centering	      
	\includegraphics[scale=0.8]{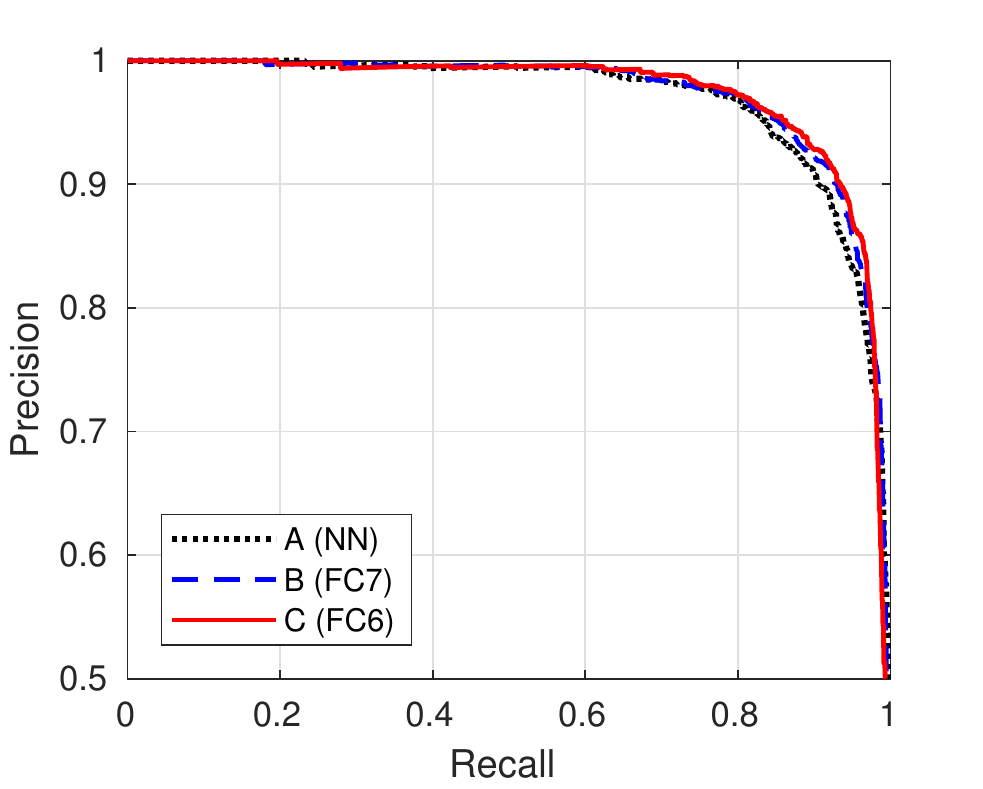}
  	\caption{\textbf{Best viewed in color.} PR curves illustrating the performance  on the validation set according to the CNN layer at which features are collected. \textit{NN} stands for prediction using solely the network softmax output layer, while \textit{FC6} and \textit{FC7} correspond to SVM classifiers trained on features collected at the first and second fully connected layers, respectively.}
  \label{fig:prNnFCx} 
\end{figure}

Results obtained for classification on the validation set are summarized in Table \ref{tab:NnFCx} and Figure \ref{fig:prNnFCx}. As Figure \ref{fig:prNnFCx} indicates, all three approaches show very similar performance. A closer inspection of Table \ref{tab:NnFCx} reveals that the SVM-based approaches slightly outperform the direct use of the neural network softmax layer both in terms of optimal $F_1$ score and AUC-PR. The performances obtained with methods B and C are very similar for both metrics. We therefore opted for method C, which uses features extracted from the earlier layer FC6 and provides slight increases in both optimal $F_1$ score and AUC-PR.

\begin{table}[!ht]\centering
  \caption{Classification performance according to the CNN layer at which features are collected - Methods A, B, C.}
  \label{tab:NnFCx}
  \begin{tabular}{@{}lcccc@{}}\toprule
	\phantom{abc}	      & \multicolumn{1}{c}{\textbf{AUC-PR}} & \multicolumn{1}{c}{$\mathbf{F_1}$} & \multicolumn{1}{c}{\textbf{Recall}} & \multicolumn{1}{c}{\textbf{Precision}}\\ \midrule[0.3pt]\midrule
    A (NN)  & $96.9\%$ & $90.6\%$  	 & $91.7\%$ 	   & $89.6\%$ \\
    B (FC7) & $97.2\%$ & $91.6\%$ 	 & $91.8\%$ 	   & $91.4\%$ \\
    C (FC6) & $97.3\%$ & $91.9\%$ 	 & $92.6\%$ 	   & $91.2\%$ \\
    \bottomrule
  \end{tabular}  
\end{table}

\subsubsection{Different types of portraits}

\begin{figure}[!ht]
	\centering	      
	\includegraphics[scale=1]{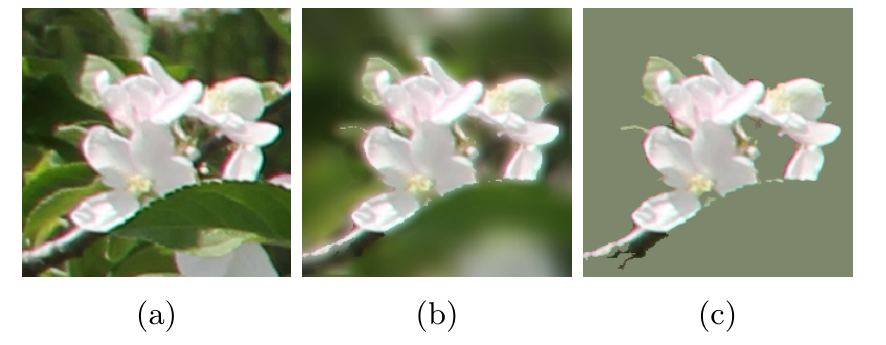}
	\caption{\textbf{Best viewed in color.} Example of the three types of portrait evaluated. a) \textit{Original}; b) Blurred background (\textit{Blur}); c) Mean padded background.}
 \label{fig:portraits}
\end{figure}

Using superpixels for region proposal computation and subsequent generation of portraits implies that our goal is to evaluate whether the superpixel itself is composed of flowers or not. In order to assess the influence of the local context surrounding the superpixel on the classification results, in addition to the approach based on replacing the region around the superpixel with the mean RGB value, two alternative approaches for portrait generation were considered. The first consists of retaining the unmodified image area surrounding the superpixel, whereas the second corresponds to blurring the background surrounding the superpixel with a low-pass filter. For all three cases, the portrait is mean-centered before being fed into the neural network. The three types of evaluated portraits are illustrated in Figure \ref{fig:portraits}.

\begin{figure}[!ht]
	\centering	      
	\includegraphics[scale=0.8]{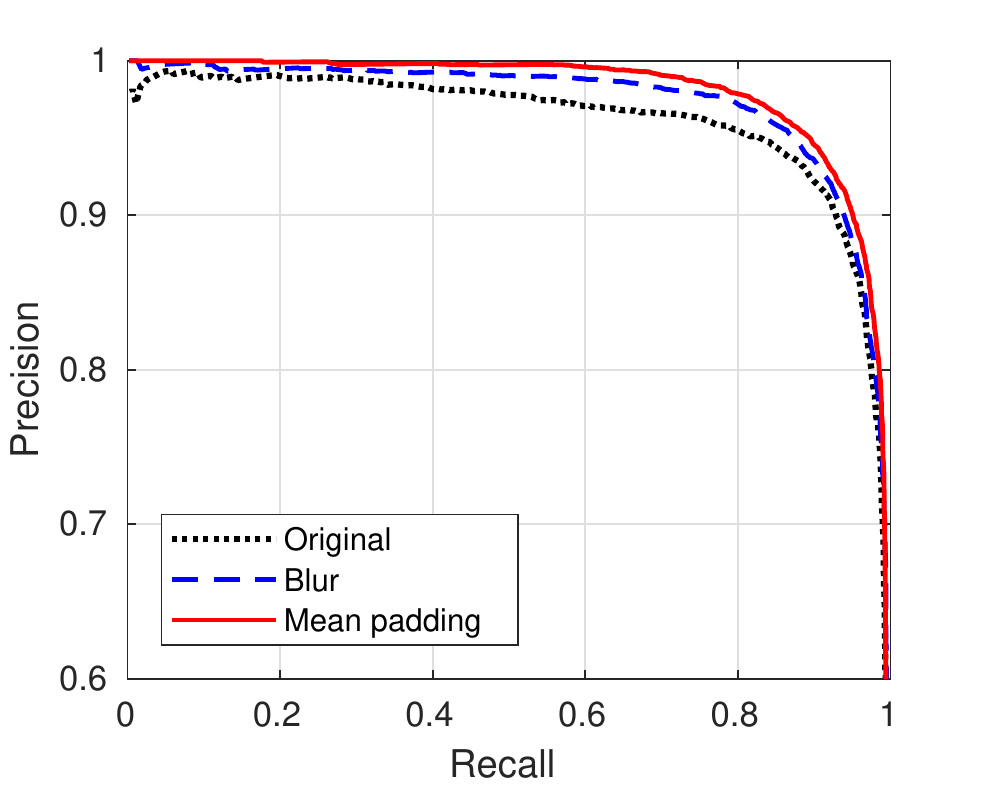}
  	\caption{\textbf{Best viewed in color.} Classification performance according to the portrait adjustment strategy. \textit{Original} stands for portraits evaluated without any further adjustment, \textit{blur} corresponds to  portraits where the background is blurred, and \textit{mean padding} denotes the strategy of padding the background with the training set mean.}
  \label{fig:prportraits} 
\end{figure}

\begin{figure}[!ht]      
	\centering	      
	\includegraphics[scale=1]{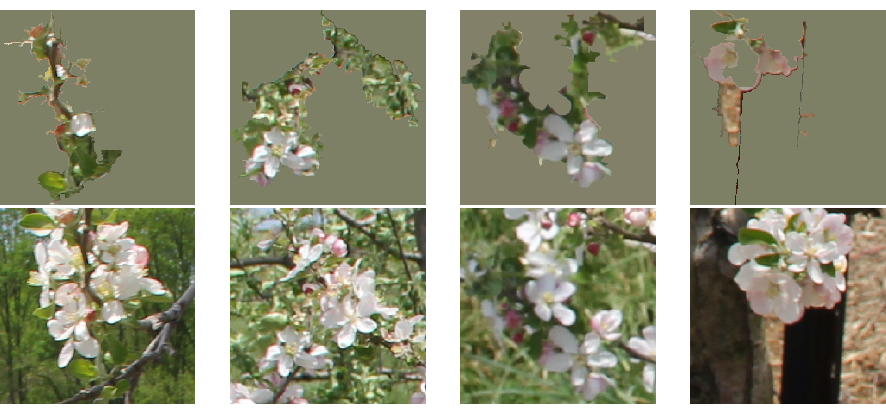}
 	\caption{\textbf{Best viewed in color.} Examples of superpixels incorrectly classified for \textit{Original} and \textit{Blur} portraits. The superpixels are shown in the top row and the bottom row shows the entire portraits enclosing the superpixels.}
   \label{fig:mistakes}
\end{figure}

Figure \ref{fig:prportraits} shows the PR curves obtained for each portrait type. The best performance is obtained with mean-padded portraits, a behavior explained by the existence of cases such as the ones illustrated in Figure \ref{fig:mistakes}. The superpixels highlighted in the images on the top row do not contain flowers in more than $50\%$ of their area and should therefore not be classified as flowers. However, these superpixels are surrounded by flowers, as depicted in the corresponding figures in the bottom row, and hence the approach of simply cropping a square region around the superpixel leads to cases in which the portrait contains a well-defined flower. As a consequence, features extracted from the CNN for the entire portrait will indicate the presence of flowers and therefore lead to high confidence false positives, which explain the non-maximal precision ratios in the upper-left part of the respective PR curve. This problem is eliminated by mean-padding the background.

\subsection{Comparison against baseline methods}
\label{sub:comparison}
The analysis in Section \ref{sub:design} above validates the design choices of our optimal CNN+SVM model described in Section \ref{sec:methods}. That is, our optimal model uses mean-padded portraits and $69$-dimensional features obtained from the FC6 layer of the CNN. In this section, we compare this optimal CNN+SVM model against the three baseline methods described in Section \ref{sub:comparison_description}. 

The parameters of all four methods were optimized using a grid search, as described in Section \ref{sec:paropt}. Optimization of the SVM hyperparameters based on $F_1$ score resulted in the following values for regularization factor ($C$) and RBF kernel bandwidth ($\gamma$): HSV+SVM ($C = 180; \gamma = 10$); CNN+SVM ($C = 30$; $\gamma = 10^{-4}$). For the HSV+Bh method, we performed an analogous grid search to optimize the standard deviation associated with the Gaussian kernel function, obtaining an optimal parameter of $\sigma = 5$. For the HSV method, we performed an extensive grid search on our training dataset to selected an optimal set of threshold values. This procedure indicated that pixels composing flowers are distributed over the entire H range of $[0,255]$, with optimal ranges of S within $[0,32]$, V within $[139,255]$, minimum size of $1,200$ pixels and maximum size of $45,000$ pixels.

Once the optimal parameters for all the classification models were determined, we evaluated the overall performance of each method using 10-fold cross-validation. All the $133,918$ samples composing the full \textit{AppleA} dataset were combined and divided into 10 folds containing $13,391$ samples each. A total of 10 iterations was performed, in which each subsample was used exactly once as validation data. 

\begin{figure}[!h]
	\centering	      
	\includegraphics[scale=0.8]{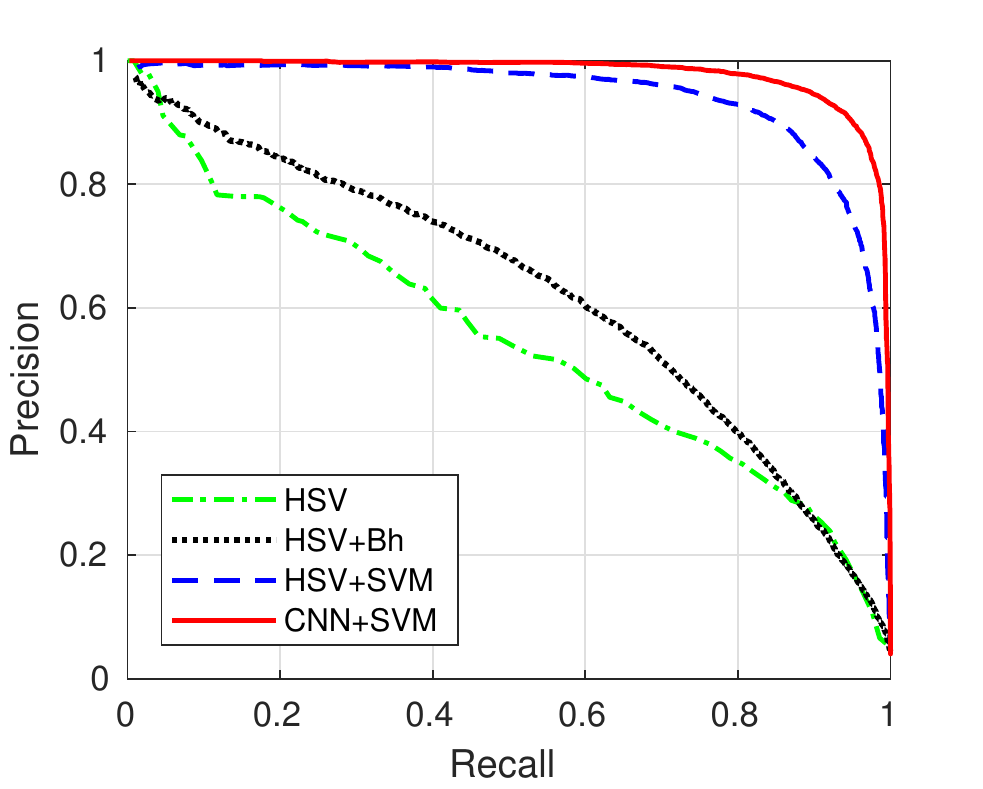}
  	\caption{\textbf{Best viewed in color.} Precision-recall (PR) curve illustrating the performance of our proposed approach (CNN+SVM) in comparison with the three baseline methods (HSV, HSV+Bh, and HSV+SVM).}
  \label{fig:PR}
\end{figure}

The final PR curves associated with each method are shown in Figure \ref{fig:PR}. Table \ref{tab:results} provides the AUC-PR for each method along with the metrics obtained for the optimal models as determined by the $F_1$ score.

\setlength{\heavyrulewidth}{0.08em}
\begin{table}[!h]\centering
  \caption{Summary of results obtained for our approach (CNN+SVM) and the three baseline methods (HSV, HSV+Bh and HSV+SVM). Best results in terms of $F_1$ and AUC-PR are shown in boldface.}
  \label{tab:results}
  \begin{tabular}{@{}lcccc@{}}\toprule
	\phantom{abc}	      & \multicolumn{1}{c}{\textbf{AUC-PR}} & \multicolumn{1}{c}{$\mathbf{F_1}$} & \multicolumn{1}{c}{\textbf{Recall}} & \multicolumn{1}{c}{\textbf{Precision}}\\ \midrule[0.3pt]\midrule
    HSV  	& $54.9\%$ & $54.1\%$  	 & $58.3\%$ 	   & $50.4\%$ \\
    HSV+Bh  & $61.6\%$ & $64.6\%$  	 & $56.9\%$ 	   & $60.5\%$ \\
    HSV+SVM & $92.9\%$ & $87.1\%$ 	 & $88.4\%$ 	   & $87.8\%$ \\
    CNN+SVM & $\mathbf{97.7\%}$ & $\mathbf{93.4\%}$ 	 & $92.0\%$ 	   & $92.7\%$ \\
    \bottomrule
  \end{tabular}
\end{table}

The HSV method, which closely replicates existing approaches for flower detection, performs poorly in terms of both recall and precision. Such low performance is expected for methods that rely solely on color information. Since these techniques do not consider morphology or higher-level context to characterize flowers, they are very sensitive to changes in illumination and to clutter. Small performance improvements are obtained using the HSV+Bh method, which replaces pixel-wise hard thresholding by HSV histogram analysis at superpixel level, thereby incorporating a limited amount of context information into its classification decisions.

\begin{figure}[!ht]
	\centering	      
	\includegraphics[scale=1]{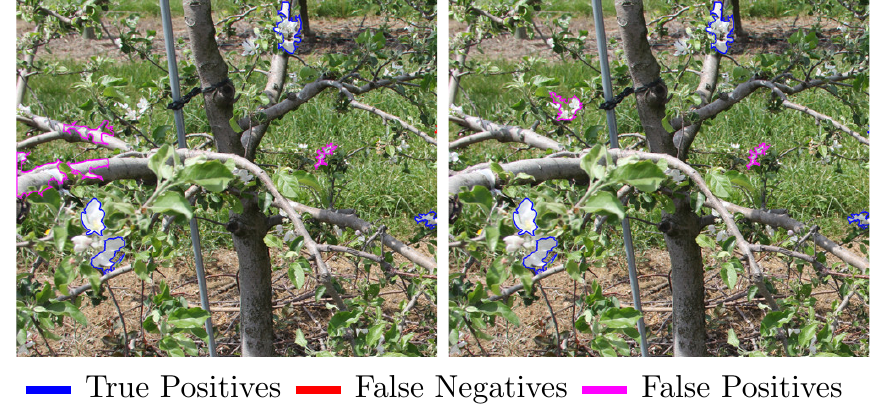}
 	\caption{\textbf{Best viewed in color.} Example of classification results obtained using (left) the baseline HSV+SVM method and (right) our proposed CNN+SVM method. Some examples of false positives generated by the HSV+SVM method that our approach correctly classifies can be seen on the branches near the left border of the image.}
   \label{fig:hsvmistake}
\end{figure}

As illustrated by the results obtained with the HSV+SVM method, the use of an SVM classifier on the same HSV color features leads to dramatic improvements in both $F_1$ and AUC-PR ratios (around $20\%$ and $30\%$, respectively). Rather than giving the same importance to all histogram regions, the SVM classifier is capable of distinguishing between poorly and highly informative features. However, as depicted in Figure \ref{fig:hsvmistake}, the precision of this method is still compromised by gross errors such as classifying parts of tree branches as flowers, since it does not take into account any morphological information.

The proposed approach (CNN+SVM) outperforms both baseline methods by extracting features using a convolutional neural network. Differently from the previous methods, the hierarchical features evaluated within the CNN take into account not only color but also morphological/spatial characteristics from each superpixel. Our results demonstrate the effectiveness of this approach, with significant improvements in both recall and precision ratios that culminate in an optimal $F_1$ score higher than $92\%$ and AUC-PR above $97\%$ for the evaluated dataset. Figure \ref{fig:imgsamples} shows examples of the final classification yielded by this method.

\subsection{Performance on additional datasets}
\begin{figure*}[h]
	\centering	      
	\includegraphics[width=\textwidth]{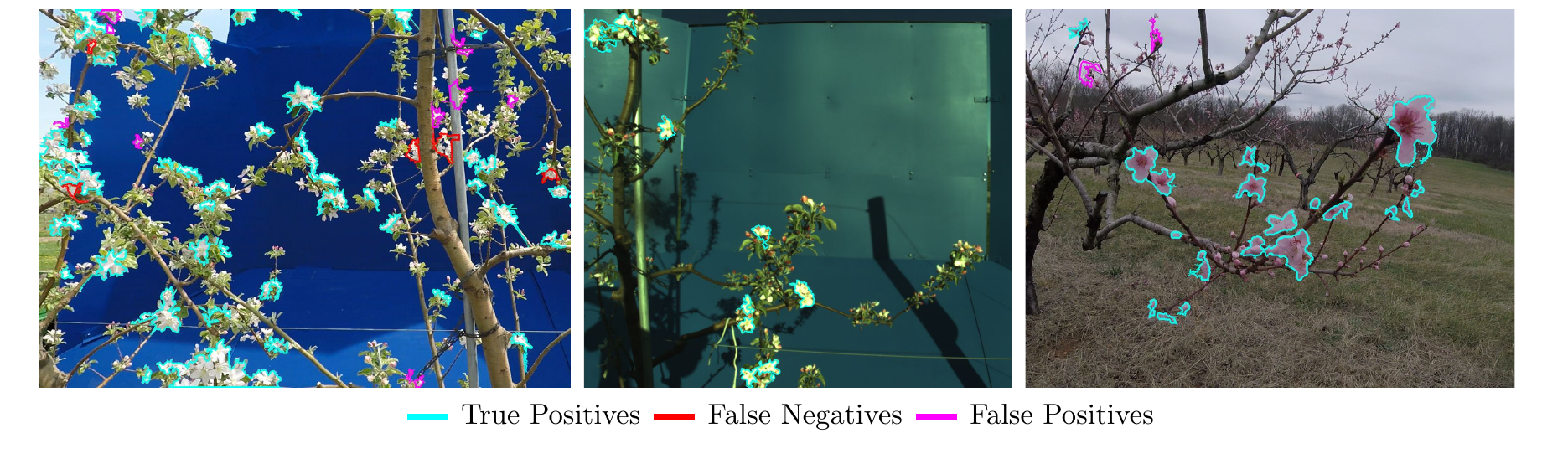}
    \caption{\textbf{Best viewed in color.} Examples of images composing the additional datasets \textit{AppleB} (left), \textit{AppleC} (middle) and \textit{Peach} (right), overlaid with the corresponding detections obtained by our method.}
    \label{fig:extrasets}
\end{figure*}

To evaluate the generalization capability of our method, we assessed its performance on three additional datasets, composed of $20$ images each and illustrated in Figure \ref{fig:extrasets}. We compare the results of our method with the performance of the best performing baseline approach (HSV+SVM). No dataset-specific adjustment of parameters is performed for our method nor for the baseline (HSV+SVM), i.e. both methods are assessed with the same optimal configuration obtained for the \textit{AppleA} dataset.

Two of the additional datasets also correspond to apple trees, but with a blue background panel positioned behind the trees to visually separate them from other rows of the orchard, a common practice in agricultural vision systems. We denote the first dataset \textit{AppleB}, which is composed of images with resolution $2704 \times 1520$ acquired using a camera model GoPro HERO5. In this dataset there is a substantial number of occlusions between branches, leaves and flowers. 

The second dataset, which we call \textit{AppleC}, is composed of images with resolution $2456 \times 2058$ acquired with a camera model JAI BB-500GE. In this dataset occlusions are less frequent but the saturation color component of the images is concentrated in a much narrower range of the spectrum than in the original \textit{AppleA} dataset. The contrast between objects such as flowers and leaves is therefore significantly lower.

The third additional dataset contains images of peach flowers (we therefore call it \textit{Peach}) with resolution $2704 \times 1520$ acquired using a camera model GoPro HERO5. Peach blossoms show a noticeable pink hue in comparison to the mostly white apple flowers composing the training dataset. Additionally, images were acquired during an overcast day, such that in comparison to the training set (\textit{AppleA}) the illumination is lower and the sky composing the background is gray instead of blue. Although the main scope of this work is on apple flower detection, we ultimately aim at a highly generalizable system that can be applied by fruit growers of different crops without the need for species-specific adjustments. In fruit orchards, each species of tree is typically constrained to specific areas. Hence, rather than differentiating between flower species, it is preferable to have a system that can distinguish between flowers and non-flower elements (e.g. leaves, branches, sky) regardless of species. Thus, this dataset represents a good evaluation of detection robustness.

\vspace{10pt}
\paragraph{Transfer learning steps} For all three additional datasets, both feature extraction and final classification were performed using the same parameters obtained by training with the \textit{AppleA} dataset, without any dataset specific fine-tuning. Our transfer learning strategy relies solely on generic pre-processing operations that approximate the characteristics of the previously unseen images to those of the training samples. 

Our first pre-processing step consists of removing the different backgrounds of the additional datasets. Whether the background is composed of a blue panel (\textit{AppleB} and \textit{AppleC}) or a gray sky (\textit{Peach}), background identification for subsequent subtraction can be performed by means of texture analysis. For each image we compute the corresponding local entropy, which is then binarized using Otsu's threshold \cite{Otsu1979} to identify low texture clusters. We then apply morphological size filtering to the binarized image and model the background as a multimodal distribution. 

To model the background, we compute the RGB-mean of the $n$ largest (in terms of number of pixels) low texture clusters to build a $n$-modal reference set. The likelihood that remaining low texture clusters belong to the background is estimated as the Euclidean distance between their means and the nearest reference in the RGB space. This metric allows differentiating between low texture components composing the background from the ones composing flowers, without any dataset specific color thresholding. For the \textit{AppleB} and \textit{Peach} datasets we adopted a bimodal distribution, where the modes correspond to the blue panel/gray sky and trunk/branches. Since the blue panel in the background of images composing the \textit{AppleC} dataset is reflective, shadows are visible and therefore we included a third mode to automatically filter these undesired elements out. Automatically determining the number of background components is part of our future work.

Afterwards, histogram equalization and histogram matching are performed on the saturation channel of each image. While equalization aims at spreading the histogram components, histogram matching consists in approximating its distribution to the characteristic form of the training set channel distribution \cite{Marques2011}. Finally, to mitigate the effects of illumination discrepancies, we subtract the difference between the mean of the value channel components in the input image and in the training set. 

\begin{figure*}[h]
	 \centering	
     \includegraphics[width=\textwidth]{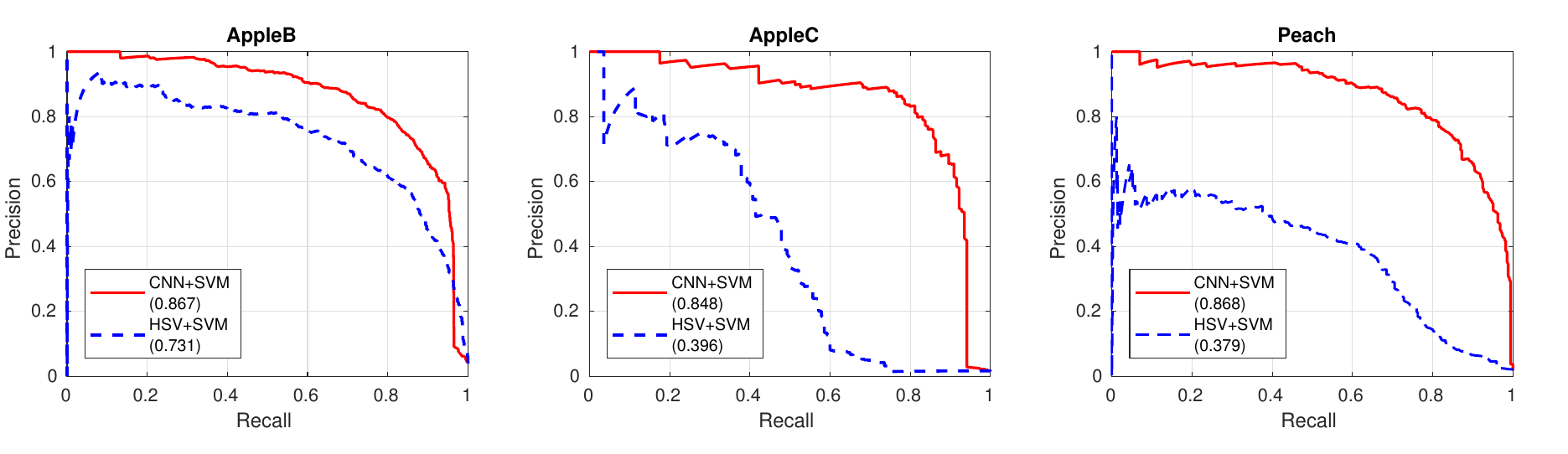}
     \caption{\textbf{Best viewed in color.} PR curves expressing the performance of our method (CNN+SVM) and the optimal baseline (HSV+SVM) approach on the three additional datasets. The AUC-PR values associated with each curve are presented within parentheses.}
    \label{fig:PRsExtraSets}
\end{figure*}

Figure \ref{fig:PRsExtraSets} shows the PR curves summarizing the performance on these datasets of our method (CNN+SVM) in comparison with the best performing baseline approach (HSV+SVM). The proposed method provides AUC-PR above $85\%$ for all datasets, significantly outperforming the baseline method. Since the HSV+SVM method relies solely on color information, its results are acceptable only for the \textit{AppleB} dataset, the one that most closely resembles the training dataset. Its performance is notably poor for the \textit{Peach} set, as this species differs to a great extent from apple flowers in terms of color. A large performance difference is also evident for the \textit{AppleC} dataset, in which flowers and leaves share more similar color components than in the training set. Table \ref{tab:generalization} shows that the proposed approach also outperforms the baseline by a large margin in terms of optimal $F_1$ score and the corresponding precision and recall values.

\begin{table}[!t]
\centering
\caption{Summary of results obtained for our approach (CNN+SVM) and the best baseline method (HSV+SVM) for the three additional datasets. Best results in terms of $F_1$ are shown in boldface.}\label{tab:generalization}
\begin{tabular}{llccc}
\hline
\textit{} &  & $\mathbf{F_1}$ & \textbf{Recall} & \textbf{Precision} \\ \hline\hline
\multirow{2}{*}{\textit{AppleB}} & HSV+SVM & $70.7\%$ & $69.8\%$ & $71.6\%$ \\
 & CNN+SVM & $\mathbf{80.2\%}$ & $81.9\%$ & $78.5\%$ \\ \hline
\multirow{2}{*}{\textit{AppleC}} & HSV+SVM & $48.6\%$ & $37.9\%$ & $68.0\%$ \\
 & CNN+SVM & $\mathbf{82.2\%}$ & $81.2\%$ & $83.3\%$ \\ \hline
\multirow{2}{*}{\textit{Peach}} & HSV+SVM & $49.0\%$ & $61.3\%$ & $40.8\%$ \\
 & CNN+SVM & $\mathbf{79.9\%}$ & $81.5\%$ & $78.3\%$ \\ \hline
\end{tabular}
\end{table}

\begin{figure}[!ht]
	\centering	
	\includegraphics[scale=1]{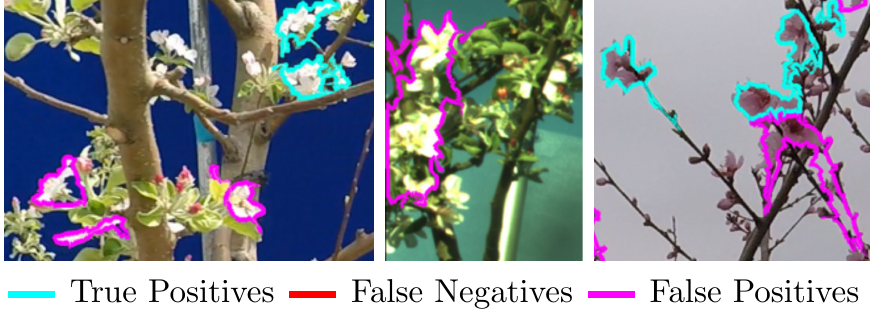}
 	\caption{\textbf{Best viewed in color.} Example of false positives caused by poor superpixel segmentation.}
   \label{fig:badsppx}
\end{figure}

Additionally, it is noteworthy that a large number of superpixels classified as false positives by our proposed approach (CNN+SVM) correspond to regions where flowers are indeed present, but compose less than $50\%$ of the corresponding superpixel total area. This is illustrated in Figure \ref{fig:badsppx}, which contains examples for the three additional datasets. In other words, the sensitivity of the feature extractor to the presence of flowers is very high and the final performance would be improved if the region proposals were more accurate.

\section{Conclusion}
\label{sec:conclusion}
In this work, we introduced a novel approach for apple flower detection, which is based on deep learning techniques that represent the state of the art for computer vision applications. In comparison with existing methods, which are mainly based solely on color analysis and have limited applicability in scenarios involving changes in illumination or occlusion levels, the hierarchical features extracted by our CNN effectively combine both color and morphological information, leading to significantly better performances for all the cases under consideration. Experiments performed on four different datasets demonstrated that the proposed CNN-based model allows accurate flower identification even in scenarios of different flower species and illumination conditions, with optimal recall and precision rates near $80\%$ even for datasets significantly dissimilar from the training sequences. 

As part of our future work, we intend to explore existing datasets and state-of-the-art models for semantic image segmentation. Particularly successful strategies consist of end-to-end architectures that, without external computation of region proposals,  generate pixel dense prediction maps for inputs with arbitrary size \cite{Shelhamer2016_FCN,Chen2017DeepLab,Lin2017RefineNet}. 

Moreover, similar to the approach proposed in \cite{Stein2016} for fruits, we intend to extend our module for flower tracking and localization based on probabilistic approaches that use the estimated motion between frames (e.g. particle filtering \cite{Mozhdehi2017}) to predict the location of flowers. To extend the applicability of our model to the detection of fruitlets as well as other flower species, we will consider additional transfer learning approaches such as data augmentation by affine transformations and the use of external datasets.

% \paragraph{Funding} We acknowledge the partial support of USDA-ARS agreement $\#58$-$8080$-$5$-$020$.

%\section*{References}
%\bibliography{refs}

{\small
\bibliographystyle{ieee}
\bibliography{techrefs_v2}
}

\end{document}